\documentclass[11pt]{article}

\usepackage[letterpaper,margin=1in]{geometry}
\usepackage{epsfig,color}
\usepackage{times}
\usepackage{amssymb,amsfonts,latexsym}
\usepackage{amsmath,amsthm}
\usepackage{enumerate,algorithmic}
\usepackage{tikz}
\usepackage{graphicx}
\usepackage[tight]{subfigure}

\theoremstyle{plain}
\newtheorem{theorem}{Theorem}
\newtheorem{corollary}{Corollary}
\newtheorem{proposition}{Proposition}
\newtheorem{lemma}{Lemma}

\newtheorem{fact}{Fact}

\theoremstyle{definition}
\newtheorem{definition}{Definition}

\newcommand{\Nat}{{\mathbb N}}
\newcommand{\ra}{\rightarrow}
\newcommand{\eps}{\epsilon}

\newcommand{\poly}{\operatorname{poly}}

\newcommand{\R}{\mathbb R}

\newcommand{\EX}{\mathit{EX}}
\newcommand{\E}{\mathbb{E}}
\newcommand{\CC}{\mathcal C}

\newcommand{\Cov}{\operatorname{Cov}}

\renewcommand{\Re}{\operatorname{Re}}
\newcommand{\dist}{\operatorname{dist}}
\renewcommand{\vec}[1]{\mbox{\boldmath$\mathrm{#1}$}}

\newcommand{\all}[1]{{\overline{#1}}}

\newcommand{\A}{\mathcal A}
\newcommand{\M}{\mathcal M}

\begin{document}

\title{Multiple Random Oracles Are Better Than One}

\author{Jan Arpe and Elchanan Mossel\\
\small U.C. Berkeley \texttt{[arpe,mossel]@stat.berkeley.edu}} 

\date{}

\maketitle

\thispagestyle{empty}

\begin{abstract}
We study the problem of learning $k$-juntas given 
access to examples drawn from a number of different product distributions. 
Thus we wish to learn a function $f : \{-1,1\}^n \to \{-1,1\}$ that depends 
on $k$ (unknown) coordinates. While the best known algorithms for the general problem of learning a $k$-junta require running time of $n^k\poly(n,2^k)$, we show that given access to $k$ different product 
distributions with biases separated by $\gamma>0$, the functions may be learned in time $\poly(n,2^k,\gamma^{-k})$. 
More generally, given access to $t \leq k$ different product distributions, the 
functions may be learned in time $n^{k/t} {\poly(n,2^k,\gamma^{-k})}$. 
Our techniques involve novel results in Fourier analysis relating Fourier 
expansions with respect to different biases and a generalization of Russo's 
formula.  
\end{abstract}


\section{Introduction}\label{sec:intro}

\subsection{Motivation}

A {\em $k$-junta} is a function $f:\{-1,1\}^n\ra\{-1,1\}$ that only depends on a subset of $k$ variables $x_{i_1},\ldots,x_{i_k}$. 
Blum and Langley~\cite{BluLan97Selection} proposed the problem of learning the class of $k$-juntas, which we refer to as the {\em junta learning probem}, as a clean and appealing model of learning in the presence of much irrelevant information. It is considered to be among the most important problems in computational learning theory to date~\cite{Blu03Learning,MoODSe04Learning}. In addition to being an interesting class in itself, the importance of learning juntas is supported by its connections to learning decision trees and DNFs, see~\cite{MoODSe04Learning}. Mossel, O'Donnell, and Servedio~\cite{MoODSe04Learning} observed that junta learning is efficiently solvable in the membership query model and in the random walk model, whereas it is provably hard in the statistical query model.
What lies in between is the uniform distribution PAC model for which~\cite{MoODSe04Learning} presented an algorithm with running time roughly $n^{0.7\cdot k}$ , being the currently best improvement upon a straightforward algorithm that runs in roughly $n^k$ steps. For general distributions, no such improvement is known. The little progress on the junta learning problem in the PAC model to date might be considered evidence of the hardness of the problem in this model. At the same time, however, no lower bounds are available, either.

Apart from devising {\em fast} learning algorithms, another goal is often to have {\em low sample complexity} (i.e., a small number of examples needed to learn). Information-theoretically, $\Theta(k\log n + 2^k)$ examples are necessary and sufficient for learning $k$-juntas on $n$ bits (\cite{BEHW89Learnability,Tur93Lower,AlmDie94Learning}). The algorithm of~\cite{MoODSe04Learning}, however, needs to draw roughly $n^{0.3k}$ examples in the worst case. 

It thus seems reasonable to ask if we can find a natural extension of the PAC learning model under fixed distributions that admits junta learning algorithms that run in time $t(k)\cdot \poly(n)$ for some function $t$ that is independent of $n$ and some polynomial that is independent of $k$. Moreover, such algorithms should ideally use $s(k)\cdot O(\log n)$ examples for some function $s$ independent of $n$.

In this paper, we propose such a model: instead of giving the learner access to only one oracle, we study the setting in which a learner has access to {\em multiple} oracles that generate examples according to different distributions. Although in this paper, we are mainly interested in learning from product distributions, we introduce the model in more generality since we believe that studying the learnability of other classes in this model, possibly under less restricted distributions, is a worthwhile goal for future research. 
In data mining and applied machine learning, researchers often depart from the assumption of having access to only one source of data in order to capture more realistic scenarios such as having multiple sources of different quality~\cite{CrKeWo05Learning,CrKeWo06Learning}, receiving partial information about tuples of examples~\cite{DZGA02Multiple}, or observing sets of different attributes for the same examples~\cite{LinYan06Discovering}. We mention three possible real-world learning scenarios in which our model can be applied: e.g., the examples could be obtained as series of measurements in certain experimental setups, so that different oracles correspond to different setups, resulting in different distributions over the instance space. Or, examples could be sampled from disjoint populations in which the distributions of attributes differ significantly. Another application comes into mind when considering data generated by a {\em mixture} of distributions. After applying algorithms to tell the distributions apart (say, from unlabeled examples)~\cite{KMR+94OnTheLearnability,VemWan02ASpectral,FeODSe05Learning}, one could use algorithms designed for the model of learning from multiple distributions to finally learn the concept under consideration.

For our results on the junta learning problem, we consider {\em $r$-biased oracles} that generate examples $(\vec x,f(\vec x))$ according to {\em $r$-biased product distributions}~$\mu_r$ on $\{-1,1\}^n$ for biases $r\in(-1,1)$. These are distributions such that every variable $x_i$ independently takes on values $-1$ and $+1$ with probability $(1-r)/2$ and $(1+r)/2$, respectively (so that $\E_{\mu_r}[x_i]=r$).

As in the setting with one uniform distribution oracle~\cite{MoODSe04Learning} (this is the case $r=0$), we show that the junta learning problem from multiple oracles reduces to the task of identifying at least one relevant variable. In general, a conceptual method to identify relevant variables is to find non-vanishing {\em Fourier coefficients} $\hat f(S,r)$, $S\subseteq[n]$, where $r$ denotes the bias of the underlying distribution. The Fourier coefficient $\hat f(S,r)$ measures the correlation between the function value $f(\vec x)$ and the function $\chi_S(\vec x,r)=\prod_{i\in r}(x_i-r)$ (see Section~\ref{ssec:Fourier} for details). Most of the literature focuses on the case $r=0$, in which $\chi_S(\vec x,0)$ reduces to the parity of the variables indexed by $S$, and $\hat f(S,0)$ is commonly denoted by $\hat f(S)$.
The point is that whenever $\hat f(S,r)\neq 0$, then all variables $x_i$ with $i\in S$ are relevant. If we pursue the search by starting with singletons $S$ and then move on to higher levels, this method takes time about $n^s$ if $\hat f(S,r)=0$ for all $S$ of size up to $s$. The question is how to proceed if such a situation occurs for some $s\in\omega(1)$. In~\cite{MoODSe04Learning} it is 
proposed to use a second approach based on the calculation of the coefficients of the polynomial representation of $f$ over the two-element field and shown a trade-off between $s$ and the degree of this polynomial. In a different direction, At\i c\i\ and Servedio~\cite{AtiSer07Quantum} enhance the uniform PAC model by a quantum subroutine to circumvent exhaustive search for non-zero Fourier coefficients. Our solution to give the learner access to several (classical) oracles can be considered as another (and maybe more realistic) alternative.

From a conceptual viewpoint, our main result shows that the junta learning problem is efficiently solvable in a {\em passive} learning model (as opposed to allowing the learner to actively ask membership queries) with {\em independent random examples} (as opposed to learning from, say, random walks, where examples are highly correlated).

\subsection{Our Results}

We solve the problem of vanishing Fourier coefficients up to high levels by considering Fourier coefficients with respect to multiple distributions: we show that if all Fourier coefficients $\hat f(S,r_i)$ of a $k$-junta $f$ vanish up to level $s$ with respect to $t$ different biases $r_1,\ldots,r_t$, then $s\cdot t<\deg(f)$, where \[\deg(f)=\max\left\{|S|\mid \hat f(S)\neq 0\right\}\leq k\] is the {\em degree} of $f$. Specifically, we prove
\begin{theorem}\label{thm:st}
Let $f:\{-1,1\}^n \ra \{-1,1\}$ be non-constant function and $s,t\in\Nat$ 
be such that $s\cdot t\geq \deg(f)$. Let $r_1,\ldots,r_t\in(-1,1)$ be arbitrary pairwise different biases. Then there exists an $i\in[t]$ and a set $S\subseteq[n]$ with $1\leq |S|\leq s$ such that $\hat f(S,r_i)\neq 0$.
\end{theorem}

Letting $s=1$ and $t=k$, Theorem~\ref{thm:st} implies that there are at most $k-1$ different biases $r$ such that all $r$-biased first-level Fourier coefficients of $f$ vanish. As a consequence, whenever a learner has access to $k$ $r$-biased oracles for $k$ pairwise distinct biases $r$, it suffices to consider, for each given bias~$r$, only coefficients $\hat f(S,r)$ at all singletons $S$ in order to find at least one relevant variable. The main technical issue we have to take care of is that Theorem~\ref{thm:st} does not rule out the possibility that $|\hat f(S,r_i)|$ could be extremely small, so that it would require a large amount of examples to tell whether a coefficient is nonzero. To take this into account, we add the requirement that the biases are well separated, i.e., have pairwise distance at least $\gamma>0$. In addition, we allow the running time to also depend on the inverse of the minimum distance of the biases to $-1$ or $1$ since the degenerate cases $r=-1$ or $r=1$ only produce the single example $(\all r,f(\all r))$, from which we cannot learn anything. Here $\all r$ denotes the vector with all $n$ entries equal to $r$. Our main learning theory application of Theorem~\ref{thm:st} (in the special case $s=1$ and $t=k$) is: 

\begin{theorem}\label{thm:level_one}
Let $-1+\alpha\leq r_1<\ldots<r_k\leq 1-\alpha$ for some $\alpha>0$ such that for all $i\in[k-1]$, $r_{i+1}-r_i\geq \gamma>0$. Then the class of $k$-juntas is exactly learnable with access to $r_i$-biased oracles, $i\in[k]$, from $m=\poly(\log n,2^k,(1/\gamma)^k,1/\alpha,\log(1/\delta))$ examples in time $\poly(m,n)$.
\end{theorem}

Theorem~\ref{thm:level_one} immediately follows from the following generalization which is based on the general case ($s\cdot t\geq k$) in Theorem~\ref{thm:st}. The trade-off between the number of $r$-biased oracles to which a learner has access and the level up to which the learner has to inspect the Fourier coefficients results in a trade-off between the number of oracles and the running time:

\begin{theorem}\label{thm:level_s}
Let $k,s,t\in\Nat$ such that $s\cdot t\geq k$ and $-1+\alpha\leq r_1<\ldots<r_t\leq 1-\alpha$ for some $\alpha>0$ such that for all $i\in[t-1]$, $r_{i+1}-r_i\geq \gamma>0$. Then  the class of $k$-juntas is exactly learnable with access to $r_i$-biased oracles, $i\in[t]$, using $m=\poly(\log n,2^k,(1/\gamma)^{k},(1/\alpha)^{s},\log (1/\delta))$ examples and running in time $n^s\cdot\poly(m,n)$. 
\end{theorem}

In other words, given access to $t$ biased oracles with biases separated by $\gamma>0$, the class of $k$-juntas is learnable in time $n^{k/t}\poly(n,2^k,\gamma^{-k})$. We should mention that we must have $\gamma\geq 2/t$ to be able to separate $t$ biases, so that $\gamma^{-k}\geq (t/2)^k$. If $t=k$, the running time is thus at least polynomial in $2^{k\log k}$.

Theorems~\ref{thm:level_one} and \ref{thm:level_s} are valid even if the biases are not known to the learner in advance. This follows since given the promise that the examples are generated according to $r$-biased product distributions, the learner can efficiently approximate these biases to within high accuracy (even from unlabeled examples) and working with such approximate biases is sufficient to recognize non-vanishing Fourier coefficients of the true biases (see Section~\ref{sec:unknown_biases}).

It is observed in~\cite{MoODSe04Learning} that except for a set of measure zero of product distributions with bias vectors $\vec r=(r_1,\ldots,r_n)\in[-1,1]^n$ (i.e., $\E[x_i]=r_i$), every $k$-junta $f$ has nonzero correlation with each of its relevant variables. They concluded that for each such vector of biases, $k$-juntas are learnable with confidence $1-\delta$ in time $\poly(2^k,n,\log(1/\delta))$. However, the correlations may become arbitrarily small, so that in order to identify nonzero correlations, these have to be approximated very precisely. As a consequence, the growth of the $\poly$ expression heavily depends on the bias vector~$\vec r$. More precisely, the running time depends on $2^{c\cdot k}$, where the constant $c$ depends on the choice of~$\vec r$.

When we restrict the product distributions to $r$-biased distributions, we can improve from a set of measure zero of exceptional bias vectors to {\em finitely} many exceptional biases: for fixed $k$ and arbitrary $n$, there are only finitely many {\em critical biases} $r\in(-1,1)$ such that there exists a $k$-junta $f$ with $\hat f(i,r)=0$ for all $i\in[n]$. 
As an application, we show
\begin{theorem}\label{thm:finite_exceptions}
Let $k\in\Nat$. Then for all but finitely many biases $r\in(-1,1)$, there exists a function $t_r:\Nat\ra\Nat$ such that $k$-juntas are exactly learnable under the $r$-biased distribution in time $t_r(k)\cdot\poly(n,\log(1/\delta))$.
\end{theorem}

Note that, unlike this rather non-constructive result, our algorithm for the ``multiple-oracles model'' works for $k$ arbitrary and unknown biased product distributions.

\subsection{Our methods}

Denote by $\E_r[f]$ the expected value of $f(\vec x)$ under the $r$-biased distribution ($r\in (-1,1)$).
Our main technical tool is a 
formula that connects the higher-order derivatives of $\E_r[f]$ 
with respect to $r$ to the Fourier weights at certain levels of the Fourier spectrum. The formula is close in spirit to Russo's well-known formula for monotone functions and generalizations thereof to arbitrary bounded functions on the hypercube.

Russo's formula~\cite{Rus81OnTheCritical} states that for {\em monotone} Boolean functions $f:\{-1,1\}^n\ra\{-1,1\}$, 
\[\frac{d}{dr}\E_r[f] = \E_r[f\cdot \sum_{i=1}^n x_i]\; .\]
More generally, the following connection between the derivative of the expectation (with respect to the bias) and correlations between the function value and the variables is known (see Grimmett~\cite[Theorem 2.34]{Gri99Percolation}): 
\begin{equation}\label{eqn:grimmett}
\frac{d}{dr}\E_r[f] = (1-r^2)^{-1}\Cov_{\vec x\sim\mu_r}[f(\vec x),\sum_{i=1}^n x_i]
\end{equation} 
(here, we have translated Grimmett's notation to our setting, and $\Cov$ denotes the covariance). Since $\hat f(i,r)=\sigma^{-1}\Cov_r[f,x_i-r]=\sigma^{-1}\Cov_r[f,x_i]$ (see Section~\ref{ssec:Fourier}), (\ref{eqn:grimmett}) can be rewritten as 
\begin{equation}\label{eqn:Russo_one}
\frac{d}{dr}\E_r[f]=(1-r^2)^{-1/2}\sum_{i=1}^n \hat f(i,r)\; .
\end{equation}

Define the {\em weight $w_s(f,r)$ of the $s$-th $r$-biased Fourier level of $f$} as the sum of all $r$-biased Fourier coefficients at level $s$, i.e., 
\[w_s(f,r)=\sum_{S\subseteq[n]:|S|=s}\hat f(S,r)\; .\]
We use the following generalization of formula~(\ref{eqn:Russo_one}) 
which we attribute to folklore (and to the best of our knowledge, has not been published before). 

\begin{theorem}[Generalization of Russo's Formula]\label{thm:russo}
Let $f:\{-1,1\}^n\ra\R$, $s\in [n]$, and $r_*\in(-1,1)$. Then
\[\frac{d^s}{dr^s}\E_r[f]\Bigr|_{r=r_*}
=\frac{s!}{(1-r_*^2)^{s/2}}\cdot w_s(f,r_*)
\; .\]
\end{theorem}
\noindent Theorem~\ref{thm:russo} follows from a similar statement for product distributions with arbitrary biases (see Proposition~\ref{prop:russo}). 
The second ingredient to prove Theorem~\ref{thm:st} is the observation that we can write
\begin{equation}\label{eqn:Erf}
\E_r[f] =\sum_{t=0}^n w_t(f,0)r^t
\end{equation}
(see Section~\ref{sec:prelim}) and that this is a polynomial in $r$ of degree at most $\deg(f)$. Moreover, this polynomial is constant (in $r$) if and only if $f$ is constant. From Theorem~\ref{thm:russo}, we obtain that if for some $r_*$, the Fourier coefficients $\hat f(S,r_*)$ vanish for all $S\subseteq[n]$ with $1\leq|S|\leq s$, then 
$(d^t/dr^t)\E_r[f]\bigr|_{r=r*}=0$ for all $t\in[s]$,
i.e., $r_*$ is an $s$-fold root of the nonzero polynomial $(d/dr)\E_r[f]$, which is of degree at most $\deg(f)-1$. Since there can be at most $(\deg(f)-1)/s$ roots of multiplicity $s$, this proves Theorem~\ref{thm:st}. To the best of our knowledge, this is the first application of Theorem~\ref{thm:russo} in theoretical computer science. Let us remark further that we obtain the following relationship between Fourier weights with respect to different measures as a consequence of Theorem~\ref{thm:russo} and Equation~(\ref{eqn:Erf}):
\begin{equation}\label{eqn:wsfr}
w_s(f,r) = (1-r^2)^{s/2}\sum_{t=s}^n \binom{t}{s}w_t(f,0)r^{t-s}\; .
\end{equation}

\subsection{Related Work}

If we restrict ourselves to subclasses of $k$-juntas $f:\{-1,1\}^n\ra\{-1,1\}$ such as monotone or symmetric juntas (i.e., juntas invariant under permutations of the relevant variables), there do exist at least partially satisfying solutions to the junta learning problem: under the uniform distribution, monotone $k$-juntas are learnable in time $\poly(n,2^k)$ from $\poly(\log n,2^k)$ examples~\cite{MoODSe04Learning} and symmetric juntas are learnable in time $n^{O(k/\log k)}\poly(n,2^k)$~\cite{KoMaMe05Learning,LMMV05OnTheFourier}. Furthermore, results for other learning more general classes under fixed product distributions have been obtained~\cite{FuJaSm91Improved,HanMan91Learning,Ser04OnLearning,BshTam96OnTheFourier}, including the polynomial time learnability of monotone $O(\log^2 n/\log^2\log n)$-juntas. Notably, also parity juntas, i.e., parities of subsets of at most $k$ variables, are efficiently learnable from product distributions (even in the presence of attribute and classification noise), with the restriction that every variable has a non-zero bias~\cite{ArpRei07Learning}. 

Recently, At{\i}c{\i}  and Servedio~\cite{AtiSer07Quantum} have studied the junta learning problem for the case that the learner has access to a uniform distribution PAC oracle plus a quantum oracle. They showed that $k$-juntas are learnable within accuracy $\eps$ from $O(\eps^{-1}k\log k)$ quantum examples and $O(2^k\log (1/\eps))$ classical (uniformly distributed) examples, both bounds being independent of $n$. Given this dramatic speed-up (which is impossible to achieve from classical queries only), we ask the more realistic question what can be done if we are given access to {\em multiple classical} oracles.

Interestingly, our results are obtained in terms of purely statistical evaluation of the given data, i.e., one can interpret the Fourier algorithm as a statistical query (SQ) algorithm with respect to several distributions. While in the original SQ model~\cite{Kea99Efficient}, in which queries are evaluated with respect to the uniform distribution on the input space, (parity) juntas are provably not efficiently learnable~\cite{BFJ+94Weakly,BshFel02OnUsing,MoODSe04Learning}, our results show that such a lower bound is not valid if queries are evaluated with respect to several distributions.

\subsection{Organization of this Paper}

We introduce all necessary prerequisites in Section~\ref{sec:prelim}. In Section~\ref{sec:russo} we present the generalization of Russo's formula. The reduction to identifying only one relevant variable is shown in Section~\ref{sec:one_variable}. In Section~\ref{sec:level_s}, we prove Theorem~\ref{thm:level_s} that addresses learnability via the $s$-th level Fourier algorithm from several oracles. Section~\ref{sec:unknown_biases} shows that the biases do not have to be known in advance. Finally, we prove Theorem~\ref{thm:finite_exceptions} in Section~\ref{sec:finite_exceptions} and propose open problems in Section~\ref{sec:open}.

\section{Preliminaries}\label{sec:prelim}

\subsection{General Notation, Juntas, and Probability Theory}

Let $\Nat=\{0,1,2,\ldots\}$, and for $n\in\Nat$, let $[n]=\{1,\ldots,n\}$. We use boldface letters such as $\vec r$, $\vec x$, and $\vec \sigma$ to denote (real) vectors of length $n$. The corresponding entries are denoted by $r_i$, $x_i$, $\sigma_i$, and so forth. For $\vec x\in\{-1,+1\}^n$ and $i\in [n]$, denote by $\vec x^{(i)}$ the vector $\vec x$ with the sign of the $i$-th entry flipped.

\begin{definition}[Relevant variables]
Let $f:\{-1,1\}^n\ra\{-1,1\}$. For $i\in[n]$, the function $f$ {\em depends} on variable $x_i$ (equivalently, $x_i$ is {\em relevant} to $f$) if there exists an $\vec x\in\{-1,1\}^n$ such that $f(\vec x^{(i)})\neq f(\vec x)$.
\end{definition}

\begin{definition}[Junta]
Let $f:\{-1,1\}^n\ra\{-1,1\}$ and $k\in[n]$. The function $f$ is a {\em $k$-junta} if it depends on at most $k$ variables.
\end{definition}

Let $x_1,\ldots,x_n$ be independent random variables taking values $-1$ and $+1$ with $\E[x_i]=r_i\in[-1,1]$. The value $r_i$ is called the {\em bias} of $x_i$. Equivalently, $\Pr[x_i=-1]=(1-r_i)/2$ and $\Pr[x_i=1]=(1+r_i)/2$. 
In this way, $\{-1,1\}^n$ is equipped with the product measure $\mu_{\vec r}$, $\vec r=(r_1,\ldots,r_n)$, given by
\[\mu_{\vec r}(\vec x)=\prod_{i=1}^n ((1+r_ix_i)/2)\]
for $\vec x\in\{-1,1\}^n$. For $f:\{-1,1\}^n\ra\R$, we denote by $\E_{\vec r}[f]$ the expectation of $f$ with respect to $\mu_{\vec r}$.
Furthermore, for $f,g:\{-1,1\}^n\ra\R$, let 
\[\Cov_{\vec r}[f,g] = \E_{\vec r}[(f-\E_{\vec r}[f])(g-\E_{\vec r}[g])] = \E_{\vec r}[f\cdot g] - \E_{\vec r}[f]\cdot \E_{\vec r}[g]\]
denote the covariance of $f$ and $g$ with respect to $\mu_{\vec r}$. Denote by $\sigma_i=(1-r_i^2)^{1/2}$ the standard deviation of $x_i$ and let $\vec \sigma = (\sigma_1,\ldots,\sigma_n)$.
We will mostly be interested in the case that all biases $r_i$ are equal. 
For $r\in[-1,1]$, let $\all r = (r,\ldots,r)$ be the vector that consists of $n$ entries that are all equal to $r$. In this case, we write $\sigma=\sigma(r)=\sqrt{1-r^2}$. 
We will frequently use that if $|r|\leq 1-\alpha$ for some $\alpha>0$, then $\sigma\geq \sqrt{\alpha}$.
The measure $\mu_{\overline r}$ is called the {\em $r$-biased product distribution}.
We also write $\mu_r$ instead of $\mu_{\all r}$, $\E_r$ instead of $\E_{\all r}$, etc.

\subsection{Learning Theory}

We introduce an extension of the classical PAC model~\cite{Val84ATheory}. Let $\CC=\bigcup_{n\in\Nat}\CC_n$ be a class of functions, where each $\CC_n$ contains some functions $f:\{-1,1\}^n\ra\{-1,1\}$ and let $\M=\bigcup_{n\in\Nat}\M_n$ be a class of {\em input distributions}, where each $\M_n$ contains distributions on $\{-1,1\}^n$. For $f\in\CC_n$ and a distribution $\mu\in\M_n$, denote by $\EX(f,\mu)$ an {\em oracle} that on request generates $\vec x\in\{-1,1\}^n$ according to $\mu$ and returns the {\em example} $(\vec x,f(\vec x))$. For $r\in[-1,1]$, we call $\EX(f,\mu_r)$ an {\em $r$-biased oracle}. Let us first review the original PAC model. The class $\CC$ is {\em PAC-learnable under distributions $\M$} if there is an algorithm $\A$ that for all $n\in\Nat$, all functions $f\in\CC_n$, and all distributions $\mu\in\M_n$ on $\{-1,1\}^n$, given $\delta,\eps>0$ and access to $\EX(f,\mu)$ but no further knowledge on $f$ and $\mu$, outputs a {\em hypothesis} $h:\{-1,1\}^n\ra\{-1,1\}$ such that with probability at least $1-\delta$ (taken over all random draws of the oracle), $\Pr_{\vec x\sim\mu}[h(\vec x)\neq f(\vec x)]\leq \eps$. If $\M_n$ is the class of {\em all} distributions on $\{-1,1\}^n$, we say that $\CC$ is {\em distribution-free PAC-learnable}. If $\M_n$ only contains the {\em uniform distribution} on $\{-1,1\}^n$, we say that $\CC$ is {\em uniform distribution PAC-learnable}. If $\M_n$ is the class of all {\em $r$-biased product distributions} $\mu_r$ on $\{-1,1\}^n$, we say that $\CC$ is {\em learnable from biased distributions}. If $\A$ even manages to output exactly $f$, (i.e., $\eps=0$), we say that $\CC$ is {\em exactly learnable}.

The performance of a learning algorithm is measured by the number of examples it requests and by its running time, both of course depending on $\delta$, $\eps$, $n$, and possibly further parameters involved in the definition of the class $\CC$.

Now we study what happens if, instead of having access to a single oracle $\EX(f,\mu)$, we admit the learning algorithm to have access to {\em multiple} (pairwise different) oracles $\EX(f,\mu_i)$, $\mu_i\in\M_n$ for $i\in[t]$. 
If we do not impose any restrictions other than being pairwise different on the distributions $\mu_i$, then the learner does not gain any power since the distributions could be arbitrarily close to each other. Thus, we allow the running time to depend on the minimum distance $\gamma$ between pairs of distributions (at this point, we leave open the choice of appropriate distance measures). 

The notion of learnability is the same as above, except that we require that the hypothesis output by a learning algorithm has to satisfy with probability at least $1-\delta$ that $\Pr_{x\sim \mu_i}[h(x)\neq f(x)]\leq \eps$ for {\em all} $i\in[t]$. In this case, we say that $\CC$ is {\em PAC-learnable from $t$ oracles under distributions $\M$ with separation $\gamma$}. 

In the following, we motivate in which variants of this very general new learning model we are interested.
Our goal is to find efficient learning algorithms for the class of $k$-juntas. More precisely, for a non-decreasing function $k:\Nat\ra\Nat$, we want to learn the class $\CC=\bigcup_{n\in\Nat}\CC_n$, where $\CC_n$ consists of all $k(n)$-juntas $f:\{-1,1\}^n\ra\{-1,1\}$.
The fastest known (exact) learning algorithm for $\CC$ in the uniform distribution PAC-learning model runs in time $n^{0.7k}\poly(n,2^k,\log(1/\delta))$~\cite{MoODSe04Learning}. Moreover, for $k\in\omega(1)$, there is {\em not any} explicit distribution $\mu$ for which $\CC$ is known to be PAC-learnable under $\mu$ in time $t(k)\cdot \poly(n,\log(1/\delta))$ with an arbitrary function $t:\Nat\ra\Nat$. It thus seems reasonable to ask if we can do any better if we are given access to more than one oracle with several {\em simple} distributions (possibly known to the learner). We will show that this is in fact the case if the distributions are biased product distributions~$\mu_{r_i}$ with well-separated biases $r_i$, even without prior knowledge on the biases (except that each $|r_i|$ should be bounded away from~$1$). Consequently, we manage to learn efficiently in the model of {\em PAC-learning from multiple biased product distributions}. The separation of biases will be reflected in the dependence of the running time on $\gamma=\min_{i\neq j} |r_i-r_j|$.

\subsection{Fourier Coefficients}\label{ssec:Fourier}

For $\vec t\in\R^n$ and $S\subseteq[n]$, define $\vec t_S=\prod_{i\in S}t_i$. In particular, for $\vec x\in\{-1,1\}^n$, $\vec x_S$ is the {\em parity} of bits in $\vec x$ indexed by $S$, and for $\vec r\in[-1,1]^n$, $\E_{\vec r}[\vec x_S]=\prod_{i\in S}\E_{\vec r}[x_i]=\vec r_S$. For $i\in[n]$ and $\vec r\in(-1,1)^n$, define $\chi_i(\vec x,\vec r)=(x_i-r_i)/\sigma_i$ and for $S\subseteq[n]$, let $\chi_S(\vec x,\vec r)=\prod_{i\in S} \chi_i(\vec x,r)$. 

The measure $\mu_{\vec r}$ induces the inner product 
\[\langle f,g\rangle_{\vec r} = \E_{\vec r}[f\cdot g]=\sum_{\vec x\in\{-1,1\}^n}\mu_{\vec r}(\vec x)f(\vec x)g(\vec x)\] 
on $\R^{\{-1,1\}^n}$. The associated norm is 
\[\|f\|_{2,\vec r} = \langle f,f\rangle_{\vec r}^{1/2} = \E_{\vec r}[f^2]^{1/2}\; .\]
The functions $\chi_S=\chi_S(\cdot,\vec r)$, $S\subseteq[n]$, form an orthonormal basis of this space with respect to $\langle \cdot,\cdot\rangle_{\vec r}$:
\[\langle \chi_S,\chi_S\rangle_{\vec r}=\E_{\vec r}[\chi_S^2]=\prod_{i\in S}\frac{\E_{\vec r}[(x_i-r_i)^2]}{\sigma_i^2}=1\; ,\] and if $i\in S\setminus T$ for some sets $S,T\subseteq[n]$, then $\E_{\vec r}[\chi_S\chi_T]=\E_{\vec r}[\chi_i]\E_{\vec r}[\chi_{S\setminus \{i\}}\chi_T]=0$ since $\E_{\vec r}[\chi_i]=0$.

We can expand any function $f:\{-1,1\}^n\ra\R$ as a linear combination of the functions $\chi_S(\cdot,\vec r)$, called the {\em Fourier expansion} of $f$ with respect to $\mu_{\vec r}$: 
\[ f=\sum_{S\subseteq[n]}\langle f,\chi_S\rangle_{\vec r} \chi_S\; ,\]
and we call $\hat f(S,\vec r)=\langle f,\chi_S\rangle_{\vec r}$ the {\em Fourier coefficient of $f$ at $S$ with respect to $\mu_{\vec r}$}. 
Note that $\chi_i(\cdot,\vec r)$ is a linear function in $x_i$ and thus $\chi_S(\cdot,\vec r)$ is a multi-linear polynomial in the variables $x_i$, $i\in S$ (of degree $|S|$). Consequently, the Fourier expansion (with respect to any $\vec r\in(-1,1)^n$) provides a representation of $f$ as a real multi-linear polynomial of degree \[\deg(f,\vec r) = \max\{k\in[n]\mid \exists S\subseteq[n]: |S|=k\wedge \hat f(S,\vec r)\neq 0\}\; .\] 
Since this degree does actually not depend on $\vec r$ (there is exactly one polynomial representation of $f$), we let $\deg(f) = \deg(f,\all 0)$.

If $S=\{i\}$ is a singleton set, we also write $\hat f(i,r)$ instead of $\hat f(\{i\},r)$. Note that for $S\neq \emptyset$, $\hat f(S,\vec r)=\Cov_{\vec r}[f,\chi_S(\cdot,\vec r)]$ since $\E_{\vec r}[\chi_S(\cdot,r)]=0$. Put in another way, 
\[\sigma^{|S|}\cdot \hat f(S,\vec r) = \Cov_{\vec r}\left[f,\prod_{i\in S}(x_i-r_i)\right]\; .\]

In case we consider the $r$-biased product measure for some $r\in(-1,1)$, we call $\hat f(S,r)=\hat f(S,\all r)$ an {\em $r$-biased Fourier coefficient}. In particular, $\hat f(\emptyset,r)=\langle f,1\rangle_r=\E_r[f]$ (again using $r$ as subscripts rather than~$\all r$). For the uniform measure $\mu_{\overline r}$ with $r=0$, the Fourier expansion of $f$ directly results in the representation of $f$ as a real multilinear polynomial in canonical form (i.e., a linear combination of monomials $\vec x_S$) since $\chi_S(\vec x,0)=\vec x_S$: $f(\vec x)=\sum_{S\subseteq[n]}\hat f(S,0)\cdot \vec x_S$.
Since for $\vec r\in[-1,1]^n$, $\E_{\vec r}[\vec x_S]=\vec r_S$, we obtain 
\begin{equation}\label{eqn:Er_poly}
\E_{\vec r}[f]= \sum_{S\subseteq[n]}\hat f(S,0)\vec r_S\; ,
\end{equation}
of which (\ref{eqn:Erf}) is the special case $\vec r=\all r$ for $r\in[-1,1]$.

The {\em weight} of the $i$-th Fourier level of $f$ with respect to $\mu_{\vec r}$ is defined to be 
\[w_i(f,{\vec r}) = \sum_{S\subseteq[n]:|S|=i}\hat f(S,\vec r)\; .\]

\begin{lemma}\label{lem:sum_w_nonzero}
Let $f:\{-1,1\}^n\ra\{-1,1\}$. If $\sum_{i=1}^n w_i(f,0)=0$, then $f$ is constant.
\end{lemma}
\begin{proof}
We have \[f(1^n)=\sum_{S\subseteq[n]}\hat f(S,0)=\sum_{i=0}^n w_i(f,0)\] is either $1$ or $-1$. Thus, if $\sum_{i=1}^n w_i(f,0)=0$, then $|\hat f(\emptyset,0)|=|w_0(f,0)|=1$, i.e., $f\equiv 1$ or $f\equiv -1$.
\end{proof}

The connection between juntas and Fourier coefficients is given by the following characterization of relevant variables:
\begin{lemma}[\cite{ArpRei07Learning,MoODSe04Learning}]
Let $f:\{-1,1\}^n\ra\{-1,1\}$, $r\in(-1,1)$, and $i\in[n]$. Then $x_i$ is relevant to $f$ if and only if there exists $S\subseteq[n]$ with $i\in S$ and $\hat f(S,r)\neq 0$.
\end{lemma}
In particular, if $\hat f(S,r)\neq 0$ for some $S\subseteq[n]$ and some $r\in(-1,1)$, then all variables $x_i$, $i\in S$, are relevant to~$f$. Thus, one way to find relevant variables is to look for non-vanishing Fourier coefficients. Furthermore, if $f$ is a $k$-junta, then $\hat f(S,r)=0$ for all $S$ with $|S|>k$, i.e., looking at coefficients up to level $k$ is sufficient for finding all relevant variables. 

\subsection{Sampling Fourier Coefficients}

To approximate biased Fourier coefficients, we will make use of the Hoeffding bound~\cite{Hoe63Probability}: 
\begin{fact}[Hoeffding bound, \cite{Hoe63Probability}]\label{lem:hoeffding}
Let $X_i$, $i\in[m]$, be mutually independent random variables taking values in $[a,b]$, $a<b$. Then for any $\eps\in[0,1]$,
\[\Pr\left[\left|\sum_{i=1}^m X_i -\sum_{i=1}^m \E[X_i]\right|\geq \eps m\right] \leq 2\exp\left(\frac{-2m\eps^2}{(b-a)^2}\right)\; .\]
\end{fact}

\begin{lemma}\label{lem:estimate_fourier}
Let $f:\{-1,1\}^n\ra\{-1,1\}$, $r\in(-1,1)$, $S\subseteq[n]$, and $\delta>0$. 
Given access to $\EX(f,r)$, we can estimate $\hat f(S,r)$ within accuracy $\eps>0$ from 
$m = \poly(2^{|S|},(1/\sigma)^{|S|},\log(1/\delta),1/\eps)$ examples in time $O(m\cdot n)$ with confidence $1-\delta$, provided that $r$ is given exactly.
\end{lemma}
\begin{proof}%[Proof of Lemma~\ref{lem:estimate_fourier}]
Draw $m=2\cdot \ln(2/\delta)\cdot(2^{|S|}/\eps)^2\cdot (1/\sigma)^{2|S|}$ examples $(\vec x^t,f(\vec x^t))$ from $\EX_r(f)$.
Define $\Delta=(\max_{x_i\in\{-1,1\}}|x_i-r|)^{|S|} = (1+|r|)^{|S|}\leq 2^{|S|}$. Let $g(\vec x)=\sigma^{|S|} f(\vec x^t)\chi_S(\vec x^t,r) \in [-\Delta,\Delta]$. 
Then, by Fact~\ref{lem:hoeffding},
\[\left|\frac{1}{m}\sum_{t=1}^m g(\vec x^t) - \sigma^{|S|}\hat f(S,r)\right|\leq \eps\sigma^{|S|}\]
with probability at least $1-\delta$. 
\end{proof}

We will deal with the case that $r$ is not exactly given in advance in Section~\ref{sec:unknown_biases}. To distinguish the cases $\hat f(S,i)=0$ and $\hat f(S,i)\neq 0$, we also need that a non-vanishing $\hat f(S,i)$ is not too small. 
For this, we will use the following (straightforward) lemma:
\begin{lemma}\label{lem:away_from_zero}
Let $h\in\mathbb R[x]$ be a polynomial of degree $d$ with leading coefficient $b$ and roots $t_1,\ldots,t_d\in\mathbb C$. 
Let $t\in\mathbb R$ and $\eps>0$ such that $|t-\Re t_i|\geq \eps$ for all $i\in[d]$. 
Then $|h(t)|\geq |b|\cdot \eps^d$.
\end{lemma}
\begin{proof} Since $h(x)=b\cdot \prod_{i\in[d]}(x-t_i)$, $|h(t)| = |b|\cdot \prod_{i\in[d]} |t-t_i|\geq |b|\cdot\prod_{i\in[d]}|t-\Re t_i|\geq |b|\cdot \eps^d$.
\end{proof}

\subsection{Derivatives}

For a $k$-fold differentiable function $f:\R^n\ra\R$ and $S=\{i_1,\ldots,i_k\}\subseteq[n]$ with pairwise different elements $i_j$, denote by 
$\frac{\partial^k}{\partial \vec x_S} f = \frac{\partial^k}{\partial x_{i_1}\ldots\partial x_{i_k}}f$
the $k$-th order partial derivative with respect to $x_{i_1},\ldots,x_{i_k}$.
\begin{lemma}\label{lem:derivatives}
Let $g\in\R[t_1,\ldots,t_n]$ be a multilinear polynomial (i.e., all exponents are at most one) and define $h\in\R[t]$ by $h(t)=g(t,\ldots,t)$. Then 
\[\frac{d^k}{dt^k}h(t) = k!\cdot\sum_{S\subseteq[n]: |S|=k}\frac{\partial^k}{\partial t_S}g(t,\ldots,t)\; .\]
\end{lemma}
\begin{proof}%[Proof of Lemma~\ref{lem:derivatives}]
The easy way to see the claim is to simply apply the chain rule. For multi-linear polynomials, though, we can as well check the claim ``by hand'':
By linearity of the construction of $h$, it suffices to check the claim for the case that $g$ is a monomial. Without loss of generality, assume that $g(t_1,\ldots,t_n) = t_1\dots t_\ell$. Let $S\subseteq[n]$ with $|S|=k$. If $S\not\subseteq [\ell]$, then clearly $(\partial^k/\partial t_S)g=0=(d^k/dt^k)h$. If $S\subseteq[\ell]$, then  $(\partial^k/\partial t_S)g(t,\ldots,t)=t^{\ell-k}$, so that 
\[k!\cdot\sum_{S\subseteq[n]: |S|=k}\frac{\partial^k}{\partial t_S}g(t\ldots,t) = k!\cdot \binom{\ell}{k}t^{\ell-k} = \frac{\ell!}{(\ell-k)!}t^{\ell-k}\; .\]
On the other hand, $h(t)=t^\ell$ and thus $(d^k/dt^k)h(t)=\ell\cdot(\ell-1)\cdot\ldots\cdot (\ell -k +1) \cdot t^{\ell-k}=\frac{\ell!}{(\ell-k)!}t^{\ell-k}$.
\end{proof}

%%%%%%%%%%%
% Russo's Formula %
%%%%%%%%%%%

\section{An Extension of Russo's Formula to General Product Distributions and Higher Order Derivatives}\label{sec:russo}

In this section, we derive our connection between derivatives or $\E_{\vec r}[f]$ and Fourier levels. In particular, we prove Theorem~\ref{thm:russo} stated in Section~\ref{sec:intro}.

\begin{proposition}\label{prop:russo}
Let $f:\{-1,1\}^n\ra\R$, $S\subseteq[n]$ with $|S|=k$, and $\vec{r^*}\in(-1,1)^n$. Then 
\[\frac{\partial^k}{\partial \vec r_S}\E_{\vec r}[f]\Bigr|_{\vec r=\vec{r^*}} = \prod_{i\in S} (1-{r^*_{i}}^2)^{-1}\cdot\Cov_{\vec{r^*}}\left[f,\prod_{i\in S}(x_{i}-r^*_{i})\right] = \vec{\sigma^*_{\mathit S}}^{-1}\cdot \hat f(S,\vec{r^*})\; .\]
\end{proposition}
\begin{proof}
Expanding $f$ with respect to $\mu_{\vec{r^*}}$, we see that 
\[\E_{\vec r}[f] = \sum_{S\subseteq[n]} \hat f(S,\vec{r^*})\E_{\vec r}[\chi_S(\cdot,\vec{r^*})] = \sum_{S\subseteq[n]}\hat f(S,\vec{r^*})\prod_{i\in S}\frac{r_i - r_i^*}{\sigma_i^*}\] 
is simply the Taylor expansion of the multi-linear polynomial $\E_{\vec r}[f]$, and the claim follows.
\end{proof}

\noindent Putting together Proposition~\ref{prop:russo} and Equation~(\ref{eqn:Er_poly}), we obtain the relationship
\begin{equation}\label{eqn:fSr}
\hat f(S,\vec r) = \vec \sigma_S\sum_{T\supseteq S}\hat f(T,0)\vec r_{T\setminus S}\; .
\end{equation}
Theorem~\ref{thm:russo} in the introduction now follows from Proposition~\ref{prop:russo} and Lemma~\ref{lem:derivatives}, and (\ref{eqn:wsfr}) is a special case of~(\ref{eqn:fSr}).

\section{Identifying One Relevant Variable Is Enough}\label{sec:one_variable}

In analogy to Proposition~6 in \cite{MoODSe04Learning}, we show that if we have an algorithm that identifies just {\em one} relevant variable of a non-constant $k$-junta $f$ using $m=\poly(\log n,2^k,(1/\alpha)^k,\log(1/\delta))$ examples from $\EX(f,r_1),\ldots,\EX(f,r_t)$ (where $\alpha>0$ bounds away the biases $r_i$ from $-1$ and $1$) in time $n^{\beta k}\poly(m,n)$, then we can construct an algorithm that identifies {\em all} relevant variables and outputs the truth table of $f$ using $m'=t\cdot \poly(\log n, 2^k,(1/\alpha)^k,\log(1/\delta))$ examples in time $n^{\beta k}\poly(m',n)$ (for the same $\beta$, but a different polynomial):
\begin{proposition}\label{prop:one_variable_is_enough}
Let $\mathcal A$ be an algorithm that, given access to $\EX(f,r_1),\ldots,\EX(f,r_t)$ for some non-constant $k$-junta $f:\{-1,1\}^n\ra\{-1,1\}$ and some $r\in(-1+\alpha,1-\alpha)$ ($\alpha>0$) and given $\delta>0$, outputs with probability at least $1-\delta$ one relevant variable of $f$ using $m=\poly(\log n,2^k,(1/\alpha)^k,\log(1/\delta))$ examples in time $n^{\beta k}\cdot\poly(m,n)$. Then there is an algorithm $\mathcal B$ that, for any $k$-junta $f:\{-1,1\}^n\ra\{-1,1\}$, given access to $\EX(f,r_1),\ldots,\EX(f,r_t)$ and $\delta>0$, outputs with probability at least $1-\delta$ {\em all} relevant variables and a truth table of $f$, using $m'=t\cdot \poly(\log n,2^k,(1/\alpha)^k,\log(1/\delta))$ examples in time $n^{\beta k}\poly(m',n)$.
\end{proposition}
\begin{proof}%[Proof of Proposition~\ref{prop:one_variable_is_enough}.]
The proposition can be proved by an adaption of the proof of Proposition in \cite{MoODSe04Learning}, so we only point to the necessary modifications of the latter. First, if $f$ is non-constant, then each output value $f(x)$ is drawn from $\EX_{r_i}(f)$ with frequency at least $(\min\{(1-r_i)/2,(1+r_i)/2\})^k\geq (\alpha/2)^k$. Thus, the check for constancy with confidence $\delta$ requires $O((2/\alpha)^k\log(1/\delta))$ examples and $\poly((2/\alpha)^k,n,\log(1/\delta))$ steps. 

Next, for restrictions $f|_\rho$ of $f$ fixing at most $k$ variables, each simulation of a draw from $\EX_{r_i}(f|_\rho)$ requires the draw of $O((2\alpha)^k\log(m/\delta))$ from $\EX_{r_i}(f)$.

Since $\A$ is run at most $k2^k$ times with confidence $1-\delta/(k2^k)$ each, it suffices to draw \[O(m(2/\alpha)^k\log(mk2^k/\delta))=m\log(m/\delta)\poly(2^k,(1/\alpha)^k)\]examples from each oracle $\EX(f,r_i)$ (note that $\A$ run on different restrictions may ask $m$ examples from different oracles). 

Finally, to read off a truth table of $f$ from the examples, $\poly((2/\alpha)^k,(1/\delta))$ examples (from any of the oracles) are again sufficient to ensure with probability $1-\delta$ that every possible assignment of the relevant variables appears in the examples. The claim follows since $m=\poly(\log n, 2^k,(1/\alpha)^k,\log(1/\delta))$.
\end{proof}

\section{Learning Relevant Variables via the \boldmath $s$-th Level Fourier Algorithm}\label{sec:level_s}

The goal of this section is to prove Theorem~\ref{thm:level_s}. For $s\in[n]$, let 
\[\mathcal R_s(f) \ = \ \bigl\{r_0\in(-1,1) \mid r_0=\Re (r)\text{ for some root }r\in\mathbb C\text{ of }\frac{d}{dr}\E_r[f]\text{ of multiplicity at least }s\bigr\}\; .\]
By Theorem~\ref{thm:russo}, $\mathcal R_s(f)$ contains all $r\in(-1,1)$ such that $w_1(f,r)=\ldots=w_s(f,r)=0$ and in particular all $r\in(-1,1)$ for which $\hat f(S,r)=0$ for all $S\subseteq[n]$ of size $1\leq |S|\leq s$.

\begin{lemma}\label{lem:threshold_level_s}
Let  $f:\{-1,1\}^n\ra\{-1,+1\}$ be a non-constant $k$-junta, $s\in[k]$, and $r\in(-1,1)$ such that $\dist(r,\mathcal R_s(f))\geq \gamma > 0$. Then there exists $S\subseteq[n]$ with $1\leq |S|\leq s$ such that 
$|\hat f(S,r)|\geq \sigma^s(\gamma/4)^k$. %\; .\] 
In particular, all variables $x_i$ with $i\in S$ are relevant.
\end{lemma}
\begin{proof}
Let $r_0=r$. Let $g(r)=\E_r[f]$. By~(\ref{eqn:Erf}) and Lemma~\ref{lem:sum_w_nonzero}, $g$ is a non-constant polynomial of degree $d=\deg(g)\leq \deg(f)\leq k$ with leading coefficient $w_d(f,0)$. Let $t\geq 1$ be minimal with $(d^t/dr^t)g |_{r=r_0}\neq 0$. 
Since $r_0\not\in\mathcal R_s(f)$, $t<s$. Let $h=(d^t/dr^t)g$. Then $h$ is a non-zero polynomial of degree $d-t\leq \deg(f)-t\leq k-t$. The highest coefficient of $h$ is $b=d\cdot(d-1)\cdot \ldots\cdot (d-t+1)\cdot w_d(f,0)$. 
By Lemma~\ref{lem:away_from_zero}, $|h(r_0)|\geq |b|\cdot \gamma^{d-t}$. Since $w_d$ is a non-zero integer multiple of $2^{-k}$, $|b|\geq \frac{d!}{(d-t)!}2^{-k}$. 
By Theorem~\ref{thm:russo}, $h(r_0)=t!\sigma^{-t}w_t(f,r_0)$, so that 
\[|w_t(f,r_0)| = (t!)^{-1}\sigma^t |h(r_0)|\geq \binom{d}{t}2^{-k}\sigma^t\gamma^{d-t}\; .\] 
Hence there exists $S\subseteq[n]$ with $|S|=t$ such that 
\[|\hat f(S)|\geq \binom{k}{t}^{-1}\binom{d}{t}2^{-k}\sigma^t\gamma^{d-t}\geq\binom{k}{d}^{-1}(\gamma/2)^k\sigma^s\geq (\gamma/4)^k\sigma^s\; .\] 
\end{proof}

For the remainder of this section, we assume that a learning algorithm has exact knowledge of all biases. However, we will show in Section~\ref{sec:unknown_biases} that this assumption is not necessary.

\begin{proposition}\label{prop:level_s}
There is an algorithm such that if $f:\{-1,1\}^n\ra\{-1,+1\}$ is a non-constant $k$-junta and $r\in(-1+\alpha,1-\alpha)$ (for some $\alpha>0$) is such that $\dist(r,\mathcal R_s(f))\geq \gamma$ for some $\gamma>0$, having access to the oracle $\EX(f,r)$, for any $\delta>0$ outputs at least one relevant variable of $f$ with probability at least $1-\delta$ using $m=\poly(\log n,2^k,(1/\gamma)^{k},(1/\alpha)^{s},\log(1/\delta))$ examples and running in time $n^s\cdot \poly(m,n)$. Furthermore, for arbitrary $r\in[-1+\alpha,1-\alpha]$, with probability at least $1-\delta$, any variable output by the algorithm is relevant.
\end{proposition}
\begin{proof}%[Proof of Proposition~\ref{prop:level_s}]
By Lemma~\ref{lem:threshold_level_s}, there exists $S\subseteq[n]$ with $1\leq |S|\leq s$ such that $|\hat f(S,r)|\geq \sigma^s(\gamma/4)^k\geq \alpha^{s/2}(\gamma/4)^k$.
Thus, it suffices to estimate all coefficients $\hat f(S,r)$, $S\subseteq[n]$ with $1\leq|S|\leq s$, within accuracy $\alpha^{s/2}(\gamma/4)^k/2$, each with confidence $1-\delta\cdot n^{-s}$, to identify (with probability at least $1-\delta$) at least one $S$ such that $\hat f(S,r)\neq 0$ with confidence $1-\delta$. This takes $\poly(2^{|S|},(1/\alpha)^{|S|},\log(n^s/\delta),(4/\gamma)^k(1/\alpha)^s)$ examples from the oracle $\EX(f,r)$ by Lemma~\ref{lem:estimate_fourier}, and we can reuse the same examples to estimate all coefficients (since we use a union bound for the confidence). Overall, the number of examples used is 
\[m=\poly(\log n,2^k,(1/\gamma)^{k},(1/\alpha)^{s},\log(1/\delta))\; .\]
The algorithm outputs all variables $x_i$ for which it finds a nonzero Fourier coefficient $\hat f(S)$ with $i\in S$. Since we have to check $\sum_{i=1}^s\binom{n}{i}= O(n^s)$ coefficients in the worst-case, the running time is bounded above by $n^s\cdot \poly(m,n)$. 

For the second part of the claim, note that if $\hat f(S,r)=0$ (and especially, if $S$ contains an index $i$ of some non-relevant variable), then the estimate for $|\hat f(S,r)|$ will with high probability be smaller than $\alpha^{s/2}(\gamma/4)^k/2$.
\end{proof}

\begin{theorem}\label{thm:level_s_one_var}
Let $s,t\in[k]$ such that $s\cdot t\geq k$, $\alpha,\gamma>0$, and $-1+\alpha\leq r_1<\ldots<r_t\leq 1-\alpha$ with $r_{j+1}-r_j\geq \gamma$ for all $j\in[t-1]$. Then there is an algorithm that, for any non-constant $k$-junta $f:\{-1,1\}^n\ra\{-1,+1\}$, given $\delta>0$ and having access to the oracles $\EX(f,r_1),\ldots,\EX(f,r_t)$, outputs a relevant variable of $f$ with probability at least $1-\delta$, using $m=\poly(\log n,2^k,(1/\gamma)^{k},(1/\alpha)^{s},\log (1/\delta))$ examples and running in time $n^s\cdot\poly(m,n)$. 
\end{theorem}
\begin{proof}
Let $h(r)=w_1(f,r)/\sigma=(d/dr)\E_r[f]$. Since $h$ is a nonzero polynomial of degree at most $\deg(f)-1\leq k-1$ and since $s\cdot t\geq k$, $h$ has less than $t$ roots of multiplicity at least $s$. Consequently, there exists $j\in[t]$ such that $\dist(r_j,\mathcal R_s(f))\geq\gamma/2$. Running the algorithm from Proposition~\ref{prop:level_s} for every single bias $r_j$, $j\in[t]$, (each time with confidence parameter $\delta/t$, reusing the same examples) yields the claim.
\end{proof}

\begin{proof}[Proof of Theorem~\ref{thm:level_s}]
Theorem~\ref{thm:level_s_one_var} shows that it is possible to identify at least one relevant variable from the claimed number of examples in time $n^s\cdot \poly(m,n)$. By Proposition~\ref{prop:one_variable_is_enough}, the claim follows.
\end{proof}

We note that since $h(r)$ is of degree at most $\deg(f)-1$, it actually suffices to have $s\cdot t\geq d$ oracles if we are given the promise that $\deg(f)\leq d$.

\section{Biases Unknown in Advance}\label{sec:unknown_biases}

The algorithms provided in Section~\ref{sec:level_s} require that all biases $r_i$ are precisely known to the learner. As one might expect, this assumption is not necessary since a learner can get good estimates of the biases from (unlabeled) random examples. The main technical issue is now to show that using good estimates $r_i'$ still leads to sufficiently close approximations of the Fourier coefficients with respect to the true biases $r_i$. For this it suffices to show that $\chi_S(\cdot,r_i)$ and $\chi_S(\cdot,r_i')$ are close in $L^2$.

\begin{lemma}\label{lem:L2_chi}
Let $\alpha,\gamma>0$, $r,r'\in(-1,1)$ such that $|r|\leq 1-\alpha$ and $|r-r'|\leq \gamma$, $S\subseteq[n]$. Then 
\[\|\chi_S(\cdot,r') - \chi_S(\cdot,r)\|_{2,r} \leq \frac{|S|+1}{\alpha^{1/2}\sigma'^{s}}\gamma\; .\]
\end{lemma}
To prove Lemma~\ref{lem:L2_chi}, we will first compute, given $\vec r,\vec r'\in(-1,1)^n$, the Fourier coefficients of $\chi_S(\cdot,\vec r')$ with respect to $\mu_{\vec r}$.
Although we only need $\vec r=\all r$ and $\vec{r'}=\all{r'}$ for our applications, we state the result for general bias vectors $\vec r$ and $\vec{r'}$ since the proof does not simplify for the special case.
\begin{lemma}\label{lem:coeff_parities}
Let $\vec r,\vec{r'}\in(-1,1)^n$ and $S,T\subseteq[n]$. Then
\[\widehat{\chi_S(\cdot,\vec{r'})}(T,\vec r) = \langle \chi_S(\cdot,\vec{r'}),\chi_T(\cdot,\vec r)\rangle_{\vec r} = 
\begin{cases}
0 & \text{if }T\not\subseteq S\\
\frac{\vec \sigma_T}{\vec{\sigma'}_S}(\vec{r}-\vec{r'})_{S\setminus T} & \text{if }T\subseteq S\; .
\end{cases}\; .\]
\end{lemma}
\begin{proof}
We have $\E_{\vec r}[\chi_i(\cdot,\vec{r'})]=(r_i-r'_i)/\sigma'_i$, $\E_{\vec r}[\chi_i(\cdot,\vec r)]=0$, and $\E_{\vec r}[\chi_i(\cdot,\vec r)\cdot\chi_i(\cdot,\vec{r'})] = \sigma_i/\sigma'_i$. 
The claim now follows from
\begin{eqnarray*}
\langle \chi_S(\cdot,\vec{r'}),\chi_T(\cdot,\vec r)\rangle_{\vec r} & = & \E_{\vec r}[\chi_S(\cdot,\vec{r'})\cdot \chi_T(\cdot,\vec r)]\\
& = & \prod_{i\in S\setminus T}\E_{\vec r}[\chi_i(\cdot,\vec{r'})]\cdot \prod_{i\in T\setminus S}\E_{\vec r}[\chi_i(\cdot,\vec r)]\cdot \prod_{i\in S\cap T}\E_{\vec r}[\chi_i(\cdot,\vec r)\cdot\chi_i(\cdot,\vec{r'})]\; .
\end{eqnarray*}
\end{proof}

Now we bound the $L^2$-norm of the difference between $\chi_S(\cdot,r)$ and $\chi_S(\cdot, r')$. Here we do restrict ourselves to $\vec r=\all r$ and $\vec{r'}=\all{r'}$ to avoid an increase in technicality:

\begin{proof}[Proof of Lemma~\ref{lem:L2_chi}]
By Parseval's equation, 
\[ \|\chi_S(\cdot,r) - \chi_S(\cdot,r')\|_{2,r}^2 = \sum_{T\subseteq[n]}\left(\widehat{\chi_S(\cdot,r')}(T,r) - \widehat{\chi_S(\cdot,r)}(T,r)\right)^2\; .\]
By Lemma~\ref{lem:coeff_parities} and since $\widehat{\chi_S(\cdot,r)}(T,r)=0$ unless $T=S$, all summands for $T\not\subseteq S$ vanish.
Furthermore, Lemma~\ref{lem:coeff_parities} states that for $T\subseteq S$, $\widehat{\chi_S(\cdot,r')}(T,r) = (\sigma^t/\sigma'^s)(r-r')^{s-t}$, where we let $s=|S|$ and $t=|T|$. Thus,
\begin{eqnarray*}
\|\chi_S(\cdot,r') - \chi_S(\cdot,r)\|_{2,r}^2 & \leq & (\widehat{\chi_S(\cdot,r')}(S,r) - 1)^2 + \sum_{T\subsetneq S}\widehat{\chi_S(\cdot,r')}(T,r)^2\\
& = & (\sigma^s/\sigma'^s - 1)^2 + \sum_{t=0}^{s-1} \binom{s}{t}\left((\sigma^t/\sigma'^s) \gamma^{s-t}\right)^2\\
& = & (\sigma')^{-2s} \left[(\sigma^s - \sigma'^s)^2 + (\sigma^2+\gamma^2)^s - \sigma^{2s}\right]\; .
\end{eqnarray*}
Now we use the following two facts:
\begin{fact}\label{fct:1}
For any $a,b\in[0,1]$ with $|b-a|\leq \rho$, $|a^s - b^s|\leq s\cdot \rho$.
\end{fact}
\begin{proof}
Let $a<b$. Then by convexity of the function $x\mapsto x^s$, $b^s \leq a^s + sb^{s-1}(b-a)\leq a^s + s\delta$.
\end{proof}
\begin{fact}\label{fct:2}
If $|r'-r|\leq \gamma$, then $|\sigma'-\sigma|\leq \gamma/\sigma$.
\end{fact}
\begin{proof}
Let $\sigma(r)=\sqrt{1-r^2}$. The derivative of $\sigma$ is $(d/dr)\sigma(r)=-\frac{r}{\sigma(r)}$. Since $\sigma$ is concave, we have that for any $\delta$ such that $r,r+\delta\in(-1,1)$, $\sigma(r+\delta)\leq \sigma(r) + (d/dr)\sigma(r)\delta = \sigma(r) - r\delta/\sigma(r)$. Since $|r|\leq 1$, the claim follows with $r'=r+\delta$, $|\delta|\leq\gamma$, $\sigma'=\sigma(r')$, and $\sigma=\sigma(r)$. 
\end{proof}

Let $\rho = \gamma \alpha^{-1/2}$. By Fact~\ref{fct:2} and since $\sigma^2=1-r^2\geq 1-r\geq \alpha$, $|\sigma' - \sigma|\leq \rho$. From Fact~\ref{fct:1}, we obtain $|\sigma'^s - \sigma^s|\leq s\rho$ and $(\sigma^2+\gamma^2)^s - (\sigma^2)^s \leq s\gamma^2$.
Consequently,
\[\sigma'^{2s}\|\chi_S(\cdot,r') - \chi_S(\cdot,r)\|_{2,r}^2 \leq (s\rho)^2 + s\gamma^2= s^2\gamma^2/\alpha + s\gamma^2\leq (s+1)^2\gamma^2/\alpha\; .\]
This proves the lemma.
\end{proof}

As a corollary, we obtain an estimate of how well $\langle f,\chi(\cdot,r')\rangle_r$ approximates $\hat f(S,r)$:
\begin{corollary}
Let $f:\{-1,1\}^n\ra\{-1,1\}$, $\gamma>0$, $r,r'\in(-1,1)$ such that $|r'-r|\leq \gamma$, and $S\subseteq[n]$. Then
\[\left|\langle f,\chi_S(\cdot,r')\rangle_r - \hat f(S,r)\right| \leq \frac{|S|+1}{\alpha^{1/2}\sigma'^{|S|}}\gamma\; .\]
\end{corollary}
\begin{proof}
By Cauchy-Schwartz,
\[\left|\langle f,\chi(\cdot,r')\rangle_r - \hat f(S,r)\right| = \left|\langle f,\chi_S(\cdot,r') - \chi_S(\cdot,r)\rangle_r\right|
\leq \|f\|_{2,r} \|\chi_S(\cdot,r') - \chi_S(\cdot,r)\|_{2,r}\; .\]
The claim follows since $\|f\|_{2,r}=1$.
\end{proof}

Next we show how to closely approximate $\hat f(S,r)$ given no a priori knowledge on $r$:
\begin{lemma}\label{lem:estimate_fourier_close_bias}
Let $f:\{-1,1\}^n\ra\{-1,1\}$, $\alpha>0$, $r\in[-1+\alpha,1-\alpha]$, $S\subseteq[n]$, and $\delta>0$. 
Given access to $\EX(f,r)$, we can estimate $\hat f(S,r)$ within accuracy $\eps$ from 
$m = \poly(2^{|S|},(1/\alpha)^{|S|},\log(1/\delta),1/\eps)$ examples in time $O(m\cdot n)$ with confidence $1-\delta$ {\em without any a priori knowledge on $r$}.
\end{lemma}
\begin{proof}%[Proof of Lemma~\ref{lem:estimate_fourier_close_bias}]
Let $\gamma=\alpha^{(|S|+1)/2}/(2(|S|+1))\leq \alpha^{1/2}\sigma'^{|S|}/(2|S|+1)$, so that, in particular, $\gamma\leq\alpha/2\leq\sigma^2/2$ (note that we may assume $|S|\geq 1$ without loss of generality). First, we approximate $r$ to within $\gamma$ by requesting $m_1=8\ln(4/\delta)/\gamma^2=\poly(|S|^2,(1/\alpha)^{|S|},\log(1/\delta))$ examples $(\vec x^t,f(\vec x^t))$ from $\EX(f,r)$ to compute $r'=(1/m_1)\sum_{t=1}^{m_1} x_i^t$. With probability at least $\delta/2$, $|r'-r|\leq \gamma$.

Now, letting 
$g(\vec x)=\sigma'^{|S|}f(\vec x^t)\chi_S(\vec x^t,r')$, $\phi=(m_2\sigma'^{|S|})^{-1}\sum_{t=1}^{m_2} g(\vec x^t)$ approximates $\langle f,\chi_S(\cdot,r')\rangle_r$ within accuracy $\eps/2$ given $m_2=\poly(2^{|S|},(1/\sigma')^{2|S|},\log(1/\delta),1/\eps)$ examples. Since 
\[\sigma'\geq\sigma-\gamma/\sigma\geq \sigma/2\geq \alpha^{1/2}/2\] implies $(1/\sigma')^{2|S|}\leq (4/\alpha)^{|S|}$,
$m_2$ is dominated by $\poly(2^{|S|},(1/\alpha)^{|S|},\log(1/\delta),1/\eps)$. 
Finally, 
\[|\phi - \hat f(S,r)| \leq |\phi - \langle f,\chi_S(\cdot,r')\rangle_r| + |\langle f,\chi_S(\cdot,r')\rangle_r -  \hat f(S,r)|\leq \eps/2 + (|S|+1)\alpha^{-1/2}\sigma'^{-|S|}\gamma\leq \eps\; .\]
The total number of examples to be drawn is $\max\{m_1,m_2\}$, which is of the order indicated in the claim.
\end{proof}

Using Lemma~\ref{lem:estimate_fourier_close_bias} in place of Lemma~\ref{lem:estimate_fourier} shows that Proposition~\ref{prop:level_s}, Theorem~\ref{thm:level_s_one_var}, and finally also Theorems~\ref{thm:level_one} and~\ref{thm:level_s} even hold if the biases $r_i$ are not known in advance (except for the bound $|r_i|\leq 1-\alpha$).

\section{Further Results and Open Problems}

\subsection{Learning in Polynomial Time for All But Finitely Many Biases}\label{sec:finite_exceptions}

We have seen that for each $k$-junta $f$, there are at most $k-1$ biases in $(-1,1)$ for which $w_1(f,r)=0$. Since for the $r$-biased product measure, $w_1(f,r)$ does not depend on {\em where} the relevant variables are hidden, it is not hard to see that there are at most $(k-1)\cdot 2^{O(k^2)}$ biases for which there exists {\em some} $k$-junta $f$ (for any $n$) with $w_1(f,r)=0$. Let us call these biases {\em critical}. 
Let $\mathcal S_{k}$ denote the set of biases $r\in(-1,1)$ such that there exists a function $t_r:\Nat\ra\Nat$ and a $k$-junta-learning algorithm that learns from $\EX(f,r)$ in time $t_r(k)\cdot \poly(n)$. Then $\mathcal S_{k}$ is exactly the complement of the critical points. This is because the minimum distance between any two distinct critical points is a function of $k$ only. This proves Theorem~\ref{thm:finite_exceptions} stated in the introduction. Consequently, for each $k$, there are only finitely many biases for which junta-learning may not be feasible in time polynomial in $n$. The next step (left for future research) is to find lower bounds on $t_r(k)$.

Generalizing to arbitrary product distributions with bias vector $\vec r\in(-1,1)^n$, we obtain that $w_1(f,\vec r)$ is zero only for a set of biases of measure zero (since it is the zero set of a non-constant multi-linear polynomial). Considering the polynomials $\sigma\hat f(i)$ separately for each $i\in[n]$, 
we recover the statement of~\cite{MoODSe04Learning} 
that $\hat f(i,\vec r)=0$ for all $i\in[n]$ 
only for a set of measure zero.

\subsection{Open Problems}\label{sec:open}

Next to the notoriously hard problem of designing more efficient algorithms for the junta learning problem under the uniform distribution, it would also constitute considerable progress to have,  for any concretely given fixed bias~$r\neq 0$, {\em some} algorithm improving over the $n^k$ bound. Note that we have shown in Section~\ref{sec:finite_exceptions} that for all but finitely many $r$, the degree-one algorithm works. However, it is not clear how to decide in general whether a given bias is critical. We believe that the relationship~(\ref{eqn:Erf}) between Fourier coefficients with respect to different biases could be useful to this end.

In a different direction, it seems worthwhile to further study our newly introduced model of learning from multiple oracles. Can we show positive results for other learning problems that appear to be hard in the classical PAC setting? In particular, is there an efficient algorithm for learning DNFs or decision trees from multiple distributions? What general conditions on the distributions are required to make efficient learning possible? As the {\em number} of oracles obviously constitutes a significant resource parameter, it is natural to ask if polynomial time learning of juntas is also possible from $o(k)$ oracles (maybe at least for important subclasses).

\subsection{Acknowledgement}
Jan Arpe is supported by the Postdoc-Program of the German 
Academic Exchange Service (DAAD) and by NSF Career Award DMS 0548249 and BSF 2004105. Elchanan Mossel is supported by 
NSF Career Award DMS 0548249, BSF 2004105 and DOD ONR grant N0014-07-1-05-06.

%\bibliographystyle{plain}
%\bibliography{abbreviations,janbib}

\begin{thebibliography}{10}

\bibitem{AlmDie94Learning}
Hussein Almuallim and Thomas~G. Dietterich.
\newblock {L}earning {B}oolean {C}oncepts in the {P}resence of {M}any
  {I}rrelevant {F}eatures.
\newblock {\em Artificial Intelligence}, 69(1-2):279--305, September 1994.

\bibitem{ArpRei07Learning}
Jan Arpe and R{\"u}diger Reischuk.
\newblock {L}earning {J}untas in the {P}resence of {N}oise.
\newblock {\em Theoretical Computer Science}, 384:2--21, 2007.

\bibitem{AtiSer07Quantum}
Alp At{\i}c{\i} and Rocco~A. Servedio.
\newblock {Q}uantum {A}lgorithms for {L}earning and {T}esting {J}untas.
\newblock {\em Quantum Inf. Process.}, 6(5):323--348, October 2007.

\bibitem{Blu03Learning}
Avrim Blum.
\newblock {L}earning a {F}unction of $r$ {R}elevant {V}ariables.
\newblock In Bernhard Sch{\"o}lkopf and Manfred~K. Warmuth, editors, {\em
  Computational Learning Theory and Kernel Machines, 16th Annual Conference on
  Computational Learning Theory and 7th Kernel Workshop, COLT/Kernel 2003,
  Washington, DC, USA, August 24-27, 2003, Proceedings}, volume 2777 of {\em
  Lecture Notes in Computer Science}, pages 731--733. Springer, 2003.

\bibitem{BFJ+94Weakly}
Avrim Blum, Merrick Furst, Jeffrey~C. Jackson, Michael Kearns, Yishay Mansour,
  and Steven Rudich.
\newblock {W}eakly {L}earning {DNF} and {C}haracterizing {S}tatistical {Q}uery
  {L}earning {U}sing {F}ourier {A}nalysis.
\newblock In {\em Proceedings of the Twenty-Sixth Annual ACM Symposium on
  Theory of Computing, 23-25 May 1994, Montr{\'e}al, Qu{\'e}bec, Canada}, pages
  253--262, 1994.

\bibitem{BluLan97Selection}
Avrim Blum and Pat Langley.
\newblock {S}election of {R}elevant {F}eatures and {E}xamples in {M}achine
  {L}earning.
\newblock {\em Artificial Intelligence}, 97(1-2):245--271, December 1997.

\bibitem{BEHW89Learnability}
Anselm Blumer, Andrzej Ehrenfeucht, David Haussler, and Manfred~K. Warmuth.
\newblock {L}earnability and the {V}apnik-{C}hervonenkis {D}imension.
\newblock {\em J. ACM}, 36(4):929--965, October 1989.

\bibitem{BshFel02OnUsing}
Nader~H. Bshouty and Vitaly Feldman.
\newblock {O}n {U}sing {E}xtended {S}tatistical {Q}ueries to {A}void
  {M}embership {Q}ueries.
\newblock {\em J. Mach. Learn. Res.}, 2(3):359--396, August 2002.

\bibitem{BshTam96OnTheFourier}
Nader~H. Bshouty and Christino Tamon.
\newblock {O}n the {F}ourier {S}pectrum of {M}onotone {F}unctions.
\newblock {\em J. ACM}, 43(4):747--770, July 1996.

\bibitem{CrKeWo05Learning}
Koby Crammer, Michael Kearns, and Jennifer Wortman.
\newblock {L}earning from {D}ata of {V}ariable {Q}uality.
\newblock In {\em Advances in Neural Information Processing Systems 18 [Neural
  Information Processing Systems, NIPS 2005, December 5-8, 2005, Vancouver,
  British Columbia, Canada]}. MIT Press, 2005.

\bibitem{CrKeWo06Learning}
Koby Crammer, Michael Kearns, and Jennifer Wortman.
\newblock {L}earning from {M}ultiple {S}ources.
\newblock In {\em Advances in Neural Information Processing Systems 19,
  Proceedings of the Twentieth Annual Conference on Neural Information
  Processing Systems, Vancouver, British Columbia, Canada, December 4-7, 2006
  (NIPS '06)}, pages 321--328. MIT Press, 2006.

\bibitem{DZGA02Multiple}
Daniel~R. Dooly, Qi~Zhang, Sally~A. Goldman, and Robert~A. Amar.
\newblock Multiple-{I}nstance {L}earning of {R}eal-{V}alued {D}ata.
\newblock {\em J. Mach. Learn. Res.}, 3:651--678, December 2002.

\bibitem{FeODSe05Learning}
Jon Feldman, Ryan O'Donnell, and Rocco~A. Servedio.
\newblock {L}earning {M}ixtures of {P}roduct {D}istributions over {D}iscrete
  {D}omains.
\newblock In {\em 46th Annual IEEE Symposium on Foundations of Computer Science
  (FOCS'05)}, pages 501--510. IEEE Press, 2005.

\bibitem{FuJaSm91Improved}
Merrick~L. Furst, Jeffrey~C. Jackson, and Sean~W. Smith.
\newblock Improved {L}earning of {$\mathit{AC}^0$} {F}unctions.
\newblock In Leslie~G. Valiant and Manfred~K. Warmuth, editors, {\em
  Proceedings of the Fourth Annual Workshop on Computational Learning Theory
  (COLT 1991), Santa Cruz, California, USA}, pages 317--325. Morgan Kaufmann,
  1991.

\bibitem{Gri99Percolation}
Geoffrey Grimmett.
\newblock {\em Percolation}.
\newblock Grundlehren Math.\ Wiss. Springer, 2nd edition, 1999.

\bibitem{HanMan91Learning}
Thomas~R. Hancock and Yishay Mansour.
\newblock {L}earning {M}onotone $k$-$\mu$ {DNF} {F}ormulas on {P}roduct
  {D}istributions.
\newblock In Leslie~G. Valiant and Manfred~K. Warmuth, editors, {\em
  Proceedings of the Fourth Annual Workshop on Computational Learning Theory
  (COLT 1991), Santa Cruz, California, USA}, pages 179--183. Morgan Kaufmann,
  1991.

\bibitem{Hoe63Probability}
Wassily Hoeffding.
\newblock {P}robability {I}nequalities for {S}ums of {B}ounded {R}andom
  {V}ariables.
\newblock {\em J. Amer. Statist. Assoc.}, 58:13--30, 1963.

\bibitem{Kea99Efficient}
Michael Kearns.
\newblock Efficient {N}oise-{T}olerant {L}earning from {S}tatistical {Q}ueries.
\newblock {\em J. ACM}, 45(6):983--1006, November 1998.

\bibitem{KMR+94OnTheLearnability}
Michael Kearns, Yishay Mansour, Dana Ron, Ronitt Rubinfeld, Robert~E. Schapire,
  and Linda Sellie.
\newblock On the {L}earnability of {D}iscrete {D}istributions.
\newblock In {\em Proceedings of the twenty-sixth annual ACM symposium on
  Theory of computing, Montreal, Quebec, Canada (STOC '94)}, pages 273--282.
  ACM Press, 1994.

\bibitem{KoMaMe05Learning}
Mihail~N. Kolountzakis, Evangelios Markakis, and Aranyak Mehta.
\newblock {L}earning {S}ymmetric {J}untas in {T}ime $n^{o(k)}$.
\newblock In {\em Workshop on Interface between Harmonic Analysis and Number
  Theory, Marseille, 2005}, 2005.
\newblock Available as Tech. Rep. at http://arxiv.org/abs/math.CO/0504246v1.

\bibitem{LinYan06Discovering}
Charles~X. Ling and Qiang Yang.
\newblock {D}iscovering {C}lassification from {D}ata of {M}ultiple {S}ources.
\newblock {\em Data Min. Knowl. Discov.}, 12(2-3):181--201, 2006.

\bibitem{LMMV05OnTheFourier}
Richard~J. Lipton, Evangelos Markakis, Aranyak Mehta, and Nisheeth~K. Vishnoi.
\newblock On the {F}ourier {S}pectrum of {S}ymmetric {B}oolean {F}unctions with
  {A}pplications to {L}earning {S}ymmetric {J}untas.
\newblock In {\em 20th Annual IEEE Conference on Computational Complexity (CCC
  '05)}, pages 112--119, 2005.

\bibitem{MoODSe04Learning}
Elchanan Mossel, Ryan~W. O'Donnell, and Rocco~A. Servedio.
\newblock Learning functions of $k$ relevant variables.
\newblock {\em J. Comput. System Sci.}, 69(3):421--434, November 2004.

\bibitem{Rus81OnTheCritical}
Lucio Russo.
\newblock {O}n the {C}ritical {P}ercolation {P}robabilities.
\newblock {\em Z.\ Wahrscheinlichkeitstheorie verw.\ Gebiete}, 56:229--237,
  1981.

\bibitem{Ser04OnLearning}
Rocco~A. Servedio.
\newblock {O}n learning monotone {DNF} under product distributions.
\newblock {\em Inform. and Comput.}, 193(1):57--74, August 2004.

\bibitem{Tur93Lower}
Gy\"orgy Tur\'an.
\newblock {L}ower {B}ounds for {PAC} {L}earning with {Q}ueries.
\newblock In Leslie~G. Valiant and Manfred~K. Warmuth, editors, {\em
  Proceedings of the Sixth Annual ACM Conference on Computational Learning
  Theory (COLT 1993), July 26-28, 1993, Santa Cruz, CA, USA}, pages 384--391.
  ACM, 1993.

\bibitem{Val84ATheory}
Leslie~G. Valiant.
\newblock {A} {T}heory of the {L}earnable.
\newblock {\em Commun. ACM}, 27(11):1134--1142, November 1984.

\bibitem{VemWan02ASpectral}
Santosh Vempala and Grant Wang.
\newblock {A} {S}pectral {A}lgorithm for {L}earning {M}ixtures of
  {D}istributions.
\newblock In {\em 43rd Symposium on Foundations of Computer Science (FOCS
  2002), 16-19 November 2002, Vancouver, BC, Canada, Proceedings}, page 113.
  IEEE Press, 2002.

\end{thebibliography}

\end{document}